\documentclass[runningheads]{llncs}

\usepackage[year=2024,ID=7709]{eccv}

\usepackage{eccvabbrv}

\usepackage{graphicx}
\usepackage{booktabs}

\usepackage[accsupp]{axessibility}  

\usepackage{algorithm}
\usepackage{algorithmic}

\usepackage{subcaption}
\usepackage{multirow}
\usepackage{xcolor}
\usepackage{float}

\usepackage[pagebackref,breaklinks,colorlinks,citecolor=eccvblue]{hyperref}

\usepackage{orcidlink}

\begin{document}

\title{Steal Now and Attack Later: Evaluating Robustness of Object Detection against Black-box Adversarial Attacks} 

\titlerunning{Steal Now and Attack Later}

\author{Erh-Chung Chen\inst{1} \and
Pin-Yu Chen\inst{2} \and
I-Hsin Chung\inst{2} \and
Che-Rung Lee\inst{1}}

\authorrunning{E. Chen et al.}

\institute{National Tsing Hua University, Taiwan \and
IBM Research, U.S.A.
}

\maketitle

\begin{abstract}
Latency attacks against object detection represent a variant of adversarial attacks that aim to inflate the inference time by generating additional ghost objects in a target image. However, generating ghost objects in the black-box scenario remains a challenge since information about these unqualified objects remains opaque. In this study, we demonstrate the feasibility of generating ghost objects in adversarial examples by extending the concept of "steal now, decrypt later" attacks. These adversarial examples, once produced, can be employed to exploit potential vulnerabilities in the AI service, giving rise to significant security concerns. The experimental results demonstrate that the proposed attack achieves successful attacks across various commonly used models and Google Vision API without any prior knowledge about the target model. Additionally, the average cost of each attack is less than \$ 1 dollars, posing a significant threat to AI security.

\end{abstract}

\section{Introduction}
Cybersecurity aims to protect computer systems and sensitive data from unexpected threats and attacks, such as CPU failures \cite{canella2019fallout} or backdoor attacks \cite{king2008designing}. With artificial intelligence (AI) rapidly revolutionizing various fields \cite{biswas2023role,chai2023stablevideo}, cybersecurity confronts new challenges, underscoring the importance of robust AI model design \cite{kaur2022trustworthy}. Adversarial attacks highlight AI model vulnerabilities, where subtle perturbations in inputs can alter predictions. Furthermore, attacks are achievable solely through predicted labels \cite{cheng2018query}, which poses a significant threat to real-world applications.

The task we focus on is evaluating the robustness of object detection, a technology that has found widespread use in various everyday applications. Object detection aims to label the positions and classes of all objects appearing in images. Commonly used deep learning-based models include faster R-CNN \cite{ren2015faster} and YOLO series \cite{Jocher_YOLO_by_Ultralytics_2023}. While remarkable performance has been made in those models, they are not immune to adversarial attacks that can produce ghost objects \cite{chen2023overload} or hide specific objects \cite{wu2020making}. Furthermore, a prior study showed that the target model can be deceived by a physical patch \cite{zolfi2021translucent}.

In order to make the attack more practical, black-box attacks have been proposed. where the attackers possess no prior knowledge about the victim model \cite{chen2017zoo,andriushchenko2019square}. Notably, traditional black-box attacks generate adversarial examples by leveraging differences in output information. During the inference phase, lower probability objects are removed in an internal step, and the information about these objects remains opaque to end-users, which prevents attackers from exploiting the target model with consistent outputs. This property makes it challenging for object detection, particularly for crafting ghost objects.

In this paper, we present a novel attack strategy, called the "steal now, attack later" approach. The core objective of this distinctive attack is to deceive the victim model into generating ghost objects under the black-box scenario such that the system cannot respond to the requests immediately or computing resources are exhausted. To accomplish the attack, we argue that the initial image should contain external objects from pre-collected data and project the perturbations onto the given constraints. This innovation opens up a new avenue for potential attacks, as adversarial examples crafted in this approach can be used to exploit vulnerabilities present in AI services. 

The follows are our main contributions in this paper:
\begin{itemize}
    \item The first paper studies the feasibility of latency attacks against object detection under the black-box scenario.
    \item This paper demonstrates that adversarial examples can be crafted by information retrieved from public datasets.
    \item Experimental results show that the proposed attack achieves successful attacks across various models, including Faster-RCNN \cite{ren2015faster}, Retinanet \cite{lin2017focal}, FCOS \cite{tian2019fcos}, DERT \cite{carion2020end} and YOLO \cite{Jocher_YOLO_by_Ultralytics_2023}, as well as public vision APIs provided by Google Cloud Platform (GCP).
    \item From the comprehensive analysis, deploying a private model locally as the most economical solution, supported by affordable costs associated with the proposed attack. This encourages attackers to invest in improving attack algorithms to exploit vulnerabilities in AI systems.

\end{itemize}

The rest of this paper is organized as follows. The background on object detection and adversarial attack is introduced in \Cref{sec:bkg}. \Cref{sec:method} describes the details of the proposed algorithm. The experimental results, ablation studies, and discussion are shown in \Cref{sec:exp}. Our conclusion is in the last section.

\section{Background}
\label{sec:bkg}
\subsection{Object Detection}
\begin{figure}[t]
\begin{minipage}[t]{.53\linewidth}
\centering
    \includegraphics[trim={3.0cm 5.8cm 3.3cm 5.8cm},clip,width=1.0\linewidth]{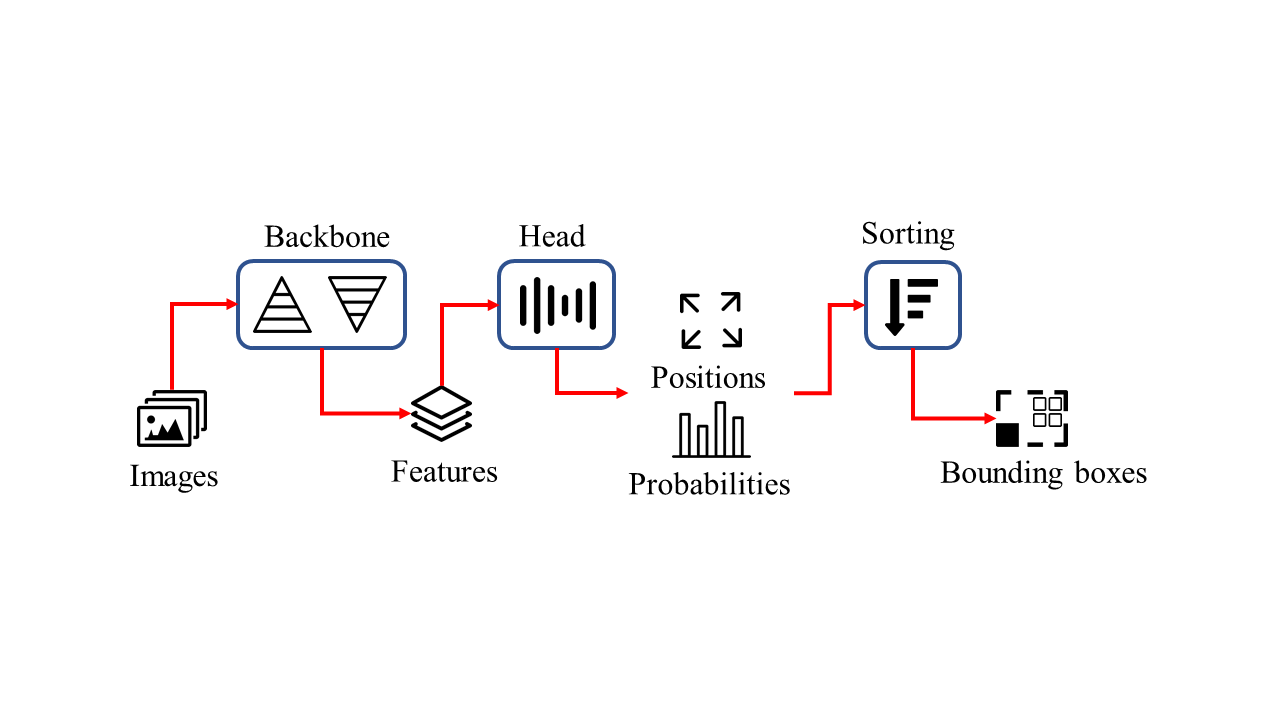}
    \caption{The execution flow of object detection.}
    \label{fig:obj_pipeline}
\end{minipage}
\begin{minipage}[t]{.45\linewidth}
\centering
    \includegraphics[trim={1.8cm 4.9cm 2.2cm 2.9cm},clip,width=1.0\linewidth]{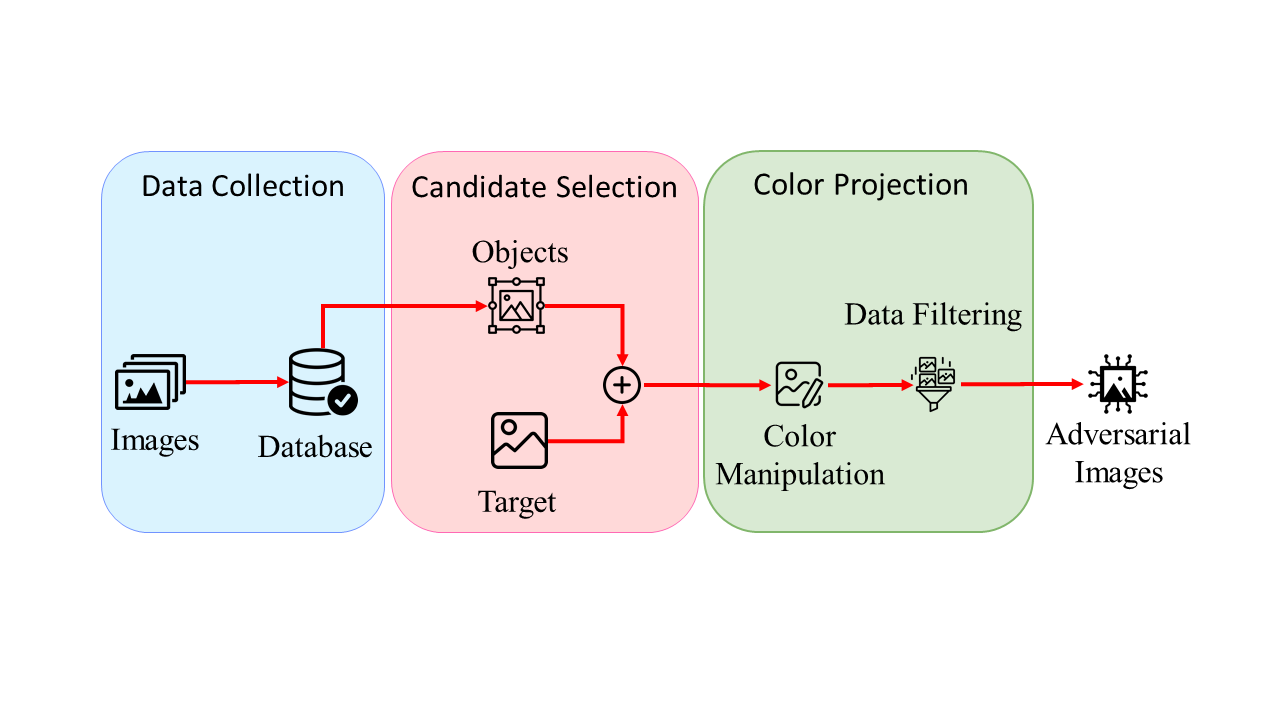}
    \caption{Attack Flow Overview.}
    \label{fig:atk_pipeline}
\end{minipage}
\end{figure}

Object detection aims to label the positions and classes of all objects within images. Despite the various models proposed in recent years, they generally follow a similar execution flow, as illustrated in \Cref{fig:obj_pipeline}. Most object detection models utilize a convolutional neural network (CNN) as the backbone, renowned for its high performance in classification tasks \cite{sultana2020review,zaidi2022survey}. The inputs are fed into the backbone model to extract features. With these features, the head model outputs locations, class probabilities, and a confidence score for all candidate objects. Based on the design of the head component, object detection models can be categorized into three types: one-stage detectors \cite{Jocher_YOLO_by_Ultralytics_2023}, two-stage detectors \cite{ren2015faster}, and transformer-based models \cite{carion2020end,beal2020toward}. Subsequently, the candidate objects are sorted, and duplicated or low-confidence objects are filtered out.



\subsection{Adversarial Attack}
The primary objective of adversarial attacks is to deceive the target models by adding malicious perturbations to the input \cite{madry2019deep,Chen_2020_ACCV}. These perturbations, however, have no impact on human judgment \cite{kaviani2022adversarial,long2022survey,hu2022adversarial}. Adversarial attacks for object detection involve different objectives. Foe example, traffic sign attacks aim to mislead automated driving systems into incorrectly recognizing traffic signs \cite{morgulis2019fooling}. Hidden attacks, on the other hand, produce physical patches that make the specific object invisible, causing automated surveillance systems to struggle with reliable detection \cite{thys2019fooling}. 

Adversarial attacks under the black-box scenario assume that any knowledge about the target model remains unknown except the output predictions \cite{mahmood2021back,liang2022large,liang2022parallel}. Common attacks involve estimating gradients through differences in output information \cite{chen2017zoo}. Score-based attacks outperform when the gradient is unreliable \cite{andriushchenko2019square}. Additionally, the existence of universal adversarial perturbations among multiple models has been demonstrated \cite{li2021universal}.

Similar to hard-label attacks \cite{cheng2018query}, object detectors internally filter out unqualified objects, rendering information about these objects opaque to end-users. This property prevents attackers from exploiting the target model with steady outputs and makes it challenging for adversarial attacks against object detection, particularly for appearing attacks \cite{cai2022zero,cai2022context} aimed at generating as many objects as possible in the target images.


 Exploring latency attacks on AI models is an evolving field. Numerous studies have shown that numerous objects appearing simultaneously in an image might inflate execution time, posing a security concern, especially for real-time object detectors \cite{shapira2023phantom,chen2023overload}. A recent study \cite{ma2023slowtrack} delved into how latency attacks can distort the camera-based autonomous driving perception pipeline. Simulation results from this study reveal alarmingly high crash rates under such attacks, underscoring the practical threats of deploying AI models in self-driving cars. However, producing latency attacks against object detection under the black-box scenario is a relatively understudied area.

\section{Methods and Algorithms}
\label{sec:method}

\subsection{Threat Model and Motivations}

In this paper, our focus is on evaluating the feasibility of the latency attack against object detection in a real-world scenario, including the total cost and time consumption of each attack as well as the success rate. The primary objective of latency attacks against object detection is to increase the processing time by introducing extra objects. We assume that the execution time for the inference system is directly proportional to the number of objects. To simplify the discussion in this paper, we focus on increasing the total number of objects. 

We assume that the latency attack is under black-box configurations since the victim models are generally deployed on the cloud, and end-users do not have any prior knowledge about the target models and the training data. Another constraint is that modification should be within a given epsilon ball for the target images to avoid the awareness of attacks. However, attackers can deploy private models locally and conduct operations as they see fit. Moreover, the utilization of publicly available datasets is deemed permissible in the pursuit of this objective.

It is important to highlight that this work primarily contributes to empirically assessing the cost of crafting adversarial examples and analyzing methods for generating ghost examples when information about the adversarial examples remains opaque. The elapsed time and potential damage caused by adversarial examples can vary depending on downstream tasks and target devices, but these aspects are beyond the scope of our study.

\subsection{Attack Flow Overview}

\begin{figure}[!t]
\centering
    \begin{subfigure}{.31\linewidth}
        \centering
        \includegraphics[width=1.0\linewidth]{fig_x_stage_0.png}
        \caption{}
        \label{fig:atk_pipeline_0}
    \end{subfigure}
    \begin{subfigure}{.31\linewidth}
        \centering
        \includegraphics[width=1.0\linewidth]{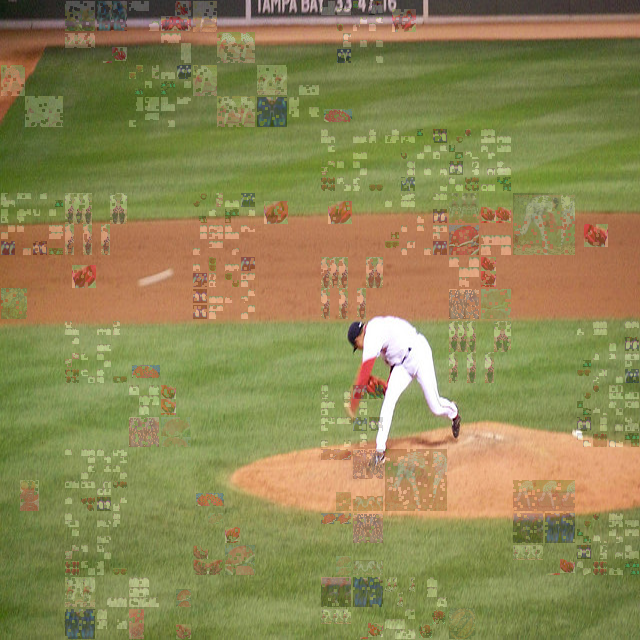}
        \caption{}
        \label{fig:atk_pipeline_1}
    \end{subfigure}
    \begin{subfigure}{.31\linewidth}
        \centering
        \includegraphics[width=1.0\linewidth]{fig_x_stage_2.png}
        \caption{}
        \label{fig:atk_pipeline_2}
    \end{subfigure}
    \begin{subfigure}{.31\linewidth}
        \centering
        \includegraphics[width=1.0\linewidth]{fig_x_stage_3.png}
        \caption{}
        \label{fig:atk_pipeline_3}
    \end{subfigure}
    \begin{subfigure}{.31\linewidth}
        \centering
        \includegraphics[width=1.0\linewidth]{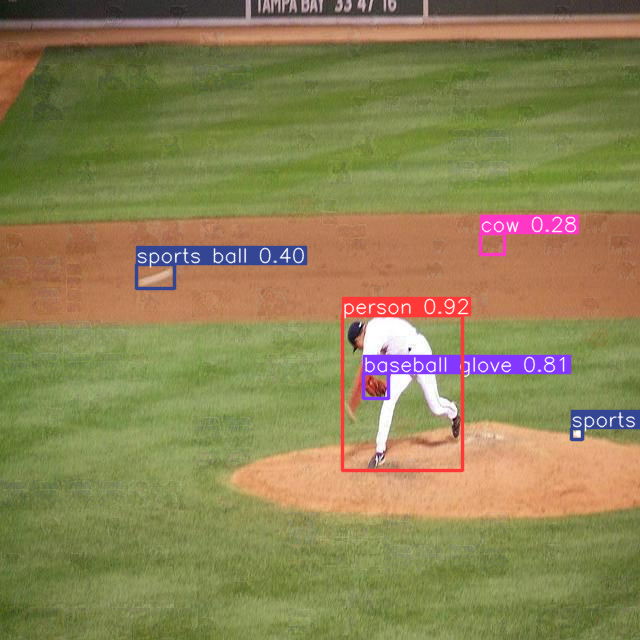}
        \caption{}
        \label{fig:atk_pipeline_0_result}
    \end{subfigure}
    \begin{subfigure}{.31\linewidth}
        \centering
        \includegraphics[width=1.0\linewidth]{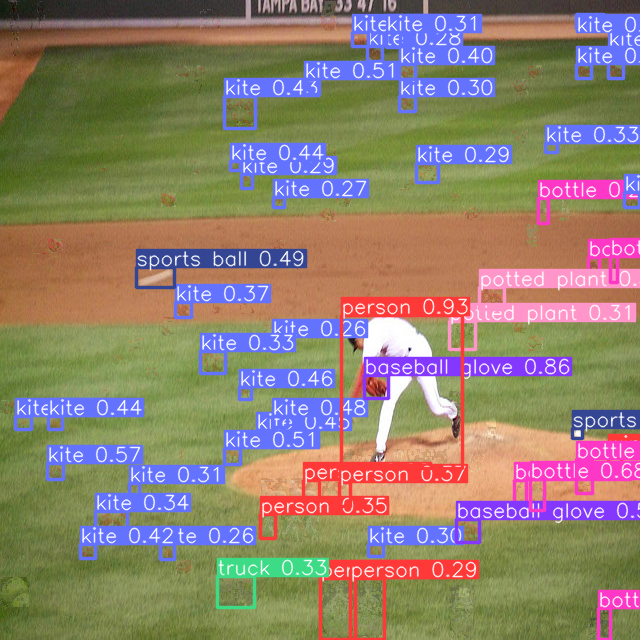}
        \caption{}
        \label{fig:atk_pipeline_3_result}
    \end{subfigure}
    \caption{Images output at each stage of the attack process. (\ref{fig:atk_pipeline_0}) is the original image; (\ref{fig:atk_pipeline_1}) shows the image with external objects; (\ref{fig:atk_pipeline_2}) displays the modified image projected onto the epsilon ball; and (\ref{fig:atk_pipeline_3}) showcases the final adversarial image. The predictions for both the original and the adversarial images are presented in (\ref{fig:atk_pipeline_0_result}) and (\ref{fig:atk_pipeline_3_result}), respectively.}
    \label{fig:atk_pipeline_output}
\end{figure}

Our proposed method adopts the spirit of the "steal now, decrypt later" attack strategy in crafting the adversarial examples. The attack flow is outlined in \Cref{fig:atk_pipeline}. Attackers gather relevant information through legitimate approaches and then store candidate objects in the database in advance. With the collected data, the attacker attempted to attach objects to the target image, thereby increasing the total number of objects predicted by the victim model. Subsequently, perturbations are projected onto the given epsilon ball within the color space and the best result is selected as the output adversarial image containing numerous ghost objects. The output images at each stage are visualized in \Cref{fig:atk_pipeline_output}.

There are three major challenges for this approach: (1) The cost of data collection from the open world and retrieval of potential objects that may exploit the victim model. (2) The determination of which objects are attached to the target image will serve as a better initial starting point. This process involves considerations of the proper sizes and suitable positions for the attached objects. (3) The formulation of an algorithm that projects the perturbations onto the given epsilon ball.

\subsection{Data Collection from Open World}
The primary objective of data collection is to gather a diverse set of objects with minimal editorial distance. We are not concerned with the accuracy of the predicted results; even false positives are deemed acceptable. Instead, our focus is on the associated costs and time investment. Therefore, any publicly available datasets, such as MS COCO \cite{lin2014microsoft}, PASCAL VOC \cite{everingham2010pascal}, or Open Images \cite{kuznetsova2020open}, which offer abundant, well-labeled, and diverse data, are eligible for the proposed attack.

Data collection involves repeatedly feeding multiple images from the collected data into object detectors to retrieve candidate objects and storing the associated information in a database. One direct approach involves sending requests to the service provider. However, there are limitations on the query rate, and each query may come with a cost, making this operation both expensive and time-consuming. On the other hand, attackers can gather information by deploying private models. By comparing the outputs of different models, attackers can effectively filter out divergent results and select consistent ones. Therefore, we believe that while collecting data by directly accessing the victim model can provide high-quality information, it is not strictly necessary.

Furthermore, we advocate the inclusion of augmented data in repositories due to the many extra benefits, despite the corresponding increased costs. Spatial crops can capture more objects at different scales or key components of objects, minimizing the influence of unimportant pixels when crafting adversarial examples. Color transformations, such as color jitter, random equalization, or random posterization, should also be considered. These operations help assess the impact of color distortion on each object. An object that becomes unrecognizable to the victim model after applying color transformations is not deemed a qualified candidate and can be filtered out in advance. These measures serve to decrease the failure rate and total number of queries when generating adversarial examples.

We argue that the collected information is not specific to just one image; it can be reused for all images as long as the victim model remains unchanged. We view data collection as a one-time cost. Additionally, we can determine if the model has been updated by comparing the output of queries for identical data at different times. Moreover, the superior performance of victim models may be leveraged to economize the cost of data collection. Specifically, the predictions from those models should have high similarity to ground-truth labels in well-known datasets. This suggests that attackers can gauge whether these public datasets furnish a comprehensive representation of the victim model's outputs through a relatively limited number of trials.

\subsection{Position-centric Candidate Selection}
\begin{algorithm}[t]
\caption{Position-centric Object Selection}
\label{alg:pos_selection}
\begin{algorithmic}[1] 
\STATE \textbf{require}: Target image $x$
\STATE \textbf{require}: Grid dimension $s_g$
\STATE \textbf{require}: Threshold $t_p$
\STATE \textbf{require}: Trials $k$
\STATE $n_w, n_h = x.\textnormal{width}/s_g, x.\textnormal{height}/s_g$
\FOR{$iter \gets 1$ \TO $k$}
    \STATE {$\mathbf{obj}=\textnormal{model.forward}(x)$}
    \label{alg:pos_selection_step1_beg}
    \FOR{$(i,j) \gets (1,1)$ \TO $(n_w, n_h)$}
        \STATE {$b \gets \textnormal{BCount}(\mathbf{obj}, i, j)$}
        \IF{$b \leq t_p$}
            \STATE {$x_{i,j} \gets \textnormal{PatchGen}(s_g)$}
        \ENDIF
    \ENDFOR
\ENDFOR

\STATE \textbf{return} $x$
\end{algorithmic}
\end{algorithm}

After gathering the necessary information, the next step involves carefully determining which object, placed in which location, would serve as a suitable candidate for creating an adversarial image. An object placed in an inappropriate position may either attract undue attention from humans or evade detection by the target model.  Estimating precise locations for each candidate object seems a potential solution. However, the object-centric approach evaluates selected objects sequentially, leaving the priority for attachment ambiguous. Consequently, different permutations of candidate objects yield varied results, with the costs of each attack directly proportional to the number of objects involved. Moreover, the inherent uncertainty necessitates the repetition of the procedure multiple times, rendering the implementation of a black-box attack impractical.

To address this, we propose an alternative position-centric algorithm inspired by the overload attack \cite{chen2023overload}. The target image is divided into a two-dimensional grid, and each grid is considered an independent component when generating adversarial examples. By evaluating the status of each grid, we determine which grid is suitable for perturbation. While some candidate objects in specific grids may not be recognized in every query, this method maximizes the total number of output objects, rendering it more cost-effective. Compared to the object-centric approach, the position-centric algorithm provides output information for all grids in a single query.

\Cref{alg:pos_selection} outlines the procedure of the position-centric algorithm, where $x$ is the target image; $s_g$ is grid dimension; $k$ is the total trials; and $t_p$ is a threshold. During each trial, the first step involves obtaining the output result predicted by the victim model for a given image. Then we traverse each grid and determine the total number of objects present in each specific grid using the function \textit{BCount}. If the number of objects is fewer than the threshold $t_p$, we assign it a new patch using \textit{PatchGen}. This function randomly selects multiple objects from the database attached to the specific grid, possibly incorporating color transformations or other operations to enrich the diversity of patches. However, if all grids undergo \textit{PatchGen}, the information about the target image is entirely eliminated. We maintain a certain probability of reverting to the original image. 

Additionally, we keep track of the status of the patch attacked on each grid, making it easier to determine which patch is optimal in subsequent steps. The information about each patch consists of the following items: the perturbed patch, the number of objects, the distance in color space, a map storing the active pixels, and a flag marking whether it is eligible. An eligible patch implies that the perturbations are within the epsilon ball and that it contains more than one object. Otherwise, a patch without any objects can be considered invalid immediately.

It is crucial to emphasize that the primary purpose of this step is not to craft an eligible adversarial image directly but rather to produce candidate patches in each grid. That is similar to the role of the initial attack of hard-label attacks \cite{cheng2018query}, where no constraints are set on the color space. A patch with a significant distance from the original image cannot be a suitable candidate because projecting the perturbations in the patch onto the epsilon ball severely distorts the underlying texture or shape of the external objects. However, a patch with perturbations outside the given epsilon ball can still serve as an initial guess.

The mechanism that decides which objects should be attached to specific grids depends on the objective. If one prefers the result of objects being uniformly distributed over the entire image, the attacker not only needs to place objects in empty regions but also reduce the confidence of objects already recognized to increase the probability that other objects will appear. For the simplicity of this work, we select the patch with the smallest editorial distance in each grid among objects whose confidence is higher than a threshold.

\subsection{Color Manipulation}
\begin{algorithm}[t]
\caption{Color Manipulation Algorithm}
\label{alg:color_projection}
\begin{algorithmic}[1] 
\STATE \textbf{require}: Target image $x$
\STATE \textbf{require}: Perturbed image $x^{adv}$
\STATE \textbf{require}: Total iterations $k_a$
\STATE \textbf{require}: Tolerance $d$
\STATE \textbf{require}: Radius of epsilon ball $\epsilon$
\label{alg:color_projection_assemble}
\FOR{$it \gets 1$ \TO $k_a$}
\label{alg:color_projection_step1_beg}
    \STATE {$\mathbf{obj}=\textnormal{model.forward}(x)$}
    \FOR{$(i,j) \gets (1,1)$ \TO $(n_w, n_h)$}
        \STATE {$b \gets \textnormal{BCount}(\mathbf{bbox}, i, j)$}
        \IF{$b < 1$}
            \STATE {$x^{adv}_{i,j} \gets \textnormal{PatchGen}(s_g)$}
        \ELSE
            \STATE {$x^{adv}_{i,j} \gets x_{i,j} + \textnormal{Projection}(x^{adv}_{i,j}-x_{i,j})$}
        \ENDIF
    \ENDFOR
    \STATE {$x^{adv}_{i,j} \gets \textnormal{Clamp}(x^{adv}_{i,j}, x_{i,j} - \epsilon d, x_{i,j} + \epsilon d)$}
\ENDFOR
\label{alg:color_projection_step1_end}

\STATE \textbf{return} $x^{adv}$
\end{algorithmic}
\end{algorithm}

The perturbed images generated through the candidate selection process described above generally fall outside the given epsilon ball. The primary objective of this step is to project these perturbed images back onto the epsilon ball while maintaining the strength of the attack. Drawing on findings from prior studies \cite{su2019one,li2023adaptive}, we believe that there are many perturbed pixels that can be dropped without influencing the final predictions. Hence, this step centers on assessing the significance of perturbations at the pixel level. Some perturbed pixels, while discernible to humans, may not give rise to ghost objects, rendering them safe to eliminate to minimize the distance. Conversely, it is imperative to identify and retain the critical perturbed pixels.

The spirit of the color manipulation is to refine the perturbations by shrinking the amplitudes of the perturbations and adjusting the average over the specific regions as well while maintaining the same predictions. The mathematical definition of \textit{Projection} is listed as follows:
\begin{equation}
\label{eq:alg_color_project}
    \mathbf{X_p} = \mathbf{M_e} \otimes F_e(\mathbf{X_p}) + (\mathbf{1} - \mathbf{M_e}) \otimes F_i(\mathbf{X_p}),
\end{equation}
where $\mathbf{X_p}$ refers to the perturbations; $\mathbf{M_e}$ is a matrix marking eligible pixels as $1$; $F_e(\cdot)$ and $F_i(\cdot)$ are two functions that operate on eligible pixels and ineligible pixels, respectively, where the ineligible pixels mean the corresponding amplitudes of the perturbations are out of the the given epsilon ball. To efficiently project the ineligible pixels onto the epsilon ball, the function $F_i$ is defined as follows:
\begin{equation}
F_i(\cdot) = s_i\mathbf{M_r} \otimes \mathbf{X_p},
\end{equation}
where $s_i$ is a scaling term and $\mathbf{M_r}$ is an random indicator matrix. When $s_i$ is set to $0$, it implies a simple dropout. On the other hand, we perform a linear transformation for eligible pixels. Specifically, $F_e(\cdot)$ can be formulated as
\begin{equation}
F_e(\cdot) = s_e\mathbf{X_p} + b_e,
\end{equation}
where $b_e$ re-centers the mean of the perturbations and $s_e$ scales the strength of the perturbations such that the distribution in the color space of the perturbed images and that of the original images are consistent. These two terms $b_e$ and $s_e$ depend on the background in the target image. For example, a camouflage perturbation should have a relatively small amplitude in low-frequency regions. 

\Cref{alg:color_projection} sketches the procedure of the color manipulation, where $d$ refers to a tolerance that allows the perturbations to exist within a larger epsilon ball during the internal steps of crafting adversarial examples. Similar to the position-centric object selection algorithm shown in \Cref{alg:pos_selection}, we traverse each grid in each iteration. If no objects are detected in a particular grid, we generate a new patch. Otherwise, we apply the function \textit{Projection} to perturbed patches. Lastly, the perturbed images are projected onto the epsilon ball with a radius of $\epsilon d$. This algorithm is reiterated with a gradual reduction in the magnitude of $d$.

While the perturbed example can cause numerous ghost objects to be predicted by the victim model, attaining the desired objective in each iteration proves elusive. The major reason is that our objective, the number of objects, is indifferentiable, rendering gradient-based black-box algorithms impractical. Therefore, we have put forth an alternative approach where the outputs for a specific $d$ are internally recorded. When obtaining worse results for smaller radii, we can revert to the previous status and repeat the procedure again to improve success rates.

\section{Experiments}
\label{sec:exp}
\subsection{Setup}
\label{sec:exp:setup}
To demonstrate the performance of the proposed attack on various models, we conducted experiments on four one-stage, one two-stage, and one transformer-based model: YOLOv8 \cite{Jocher_YOLO_by_Ultralytics_2023}, Retinanet \cite{lin2017focal}, FCOS \cite{tian2019fcos}, SSD300 \cite{liu2016ssd}, Faster-RCNN \cite{ren2015faster}, and DERT \cite{carion2020end}. We also performed the proposed algorithm on Azure vision and GCP vision API as well. We randomly selected 100 images from the validation set of MS COCO 2017 dataset \cite{lin2014microsoft} and Open Images \cite{kuznetsova2020open} as the target images. The input dimensions are fixed to $(640, 640)$ in both the data collection and inference phases. Additionally, we include ablation studies which cover various aspects including, data collection from different datasets or models and cost analysis. The predictions from different models are  visualized in Appendix A.

To the best of our knowledge, this is the first paper that focuses on appearing attacks under the black-box scenario. There is no general metric to evaluate performance. A direct comparison of total objects caused by adversarial images among models should be avoided since the maximum number of objects output by different models is not a fixed number. In this paper, we project the perturbation onto L-inf norm and define an attack as successful if the object increment caused by the adversarial example is greater than 20. 

\subsection{Performance Evaluation}
\label{sec:exp:perf_eval}

\begin{table}[t]
\centering
\caption{Attack success rates (ASR) across various models under different radii of $\epsilon$ with patches are collected from different datasets.}
\label{table:ASR_model_eps}
\begin{tabular}{c|c|c|c|c|c|c|c|c}
Dataset  & \multicolumn{4}{c|}{MS COCO} & \multicolumn{4}{|c}{Open Images} \\
\hline
\hline
Model    & $\epsilon=8$ & $\epsilon=16$ & $\epsilon=24$ & $\epsilon=32$ & $\epsilon=8$ & $\epsilon=16$ & $\epsilon=24$ & $\epsilon=32$\\
\hline
DERT            &  1\% &  7\% & 13\% & 29\%  &  1\% &  6\% & 15\% & 28\% \\
Faster-RCNN     &  2\% & 10\% & 21\% & 35\%  &  1\% &  8\% & 24\% & 36\% \\
Retinanet       &  3\% &  9\% & 22\% & 37\%  &  5\% &  8\% & 28\% & 39\% \\
FCOS            &  5\% & 12\% & 29\% & 49\%  &  4\% &  9\% & 31\% & 52\% \\
YOLOv8          &  8\% & 15\% & 42\% & 81\%  &  9\% & 13\% & 45\% & 83\% \\
\hline
SSD300          &  0\% &  0\% &  0\% &  0\%  &  0\% &  0\% &  0\% &  0\% \\
\hline
\hline
Azure           &  0\% &  0\% &  0\% &  0\%  &  0\% &  0\% &  0\% &  0\% \\
GCP             &  0\% &  0\% & 11\% & 15\%  &  0\% &  0\% & 12\% & 18\% \\
\end{tabular}

\end{table}

This experiment evaluates the attack success rates (ASR) across various models under different epsilon radii. ASR is the ratio of the number of successful attacks over the total number of attacks.  In this experiment, the maximum of query times during the attacking iteration is set to 4,000 and we collected from MS COCO and Open Images datasets. The experimental results are shown in \Cref{table:ASR_model_eps}.


As can be seen, the ASR sharply declines as the radius of the epsilon ball is reduced. This phenomenon occurs because the tiny perturbations fail to generate significant output signals, and the proposed attack is a stochastic process, making some false signals during the internal steps. Nevertheless, these results demonstrate that the attack achieves its highest ASR on the YOLOv8 model, and it is also effective on other models, with the exception of SSD300. The findings suggest the feasibility of attacking a public model, but the ASR is not notably high due to the lack of model-specific information.

ASR performance on DERT is the second worst. A possible reason is that we assume no inter-grid influence in this paper. This assumption does not align with the attention mechanism found in transformer-based models, which can dynamically adjust the importance of global input tokens, making objects have different ranges of influence in space.

Remarkably, the proposed attack encounters a definitive failure when applied to the SSD3000 model. The main issue stems from our objective design preference, which favors patches with multiple objects, indicating that each object is relatively small in size. This approach tends to divide the large object into multiple small objects during the iteration process of color manipulation. However, a previous study has demonstrated that the SSD model struggles with accurately identifying small objects \cite{huang2017speed}. Consequently, all objects vanish during these iterations. Similarly, Azure has a limitation: objects smaller than 5 \% of the image size are eliminated internally. This implies that, by definition, the attack is destined to fail.

\begin{figure}[t]
\begin{minipage}[t]{.48\linewidth}
\centering
    \begin{subfigure}{.48\linewidth}
        \centering
        \includegraphics[width=1.0\linewidth]{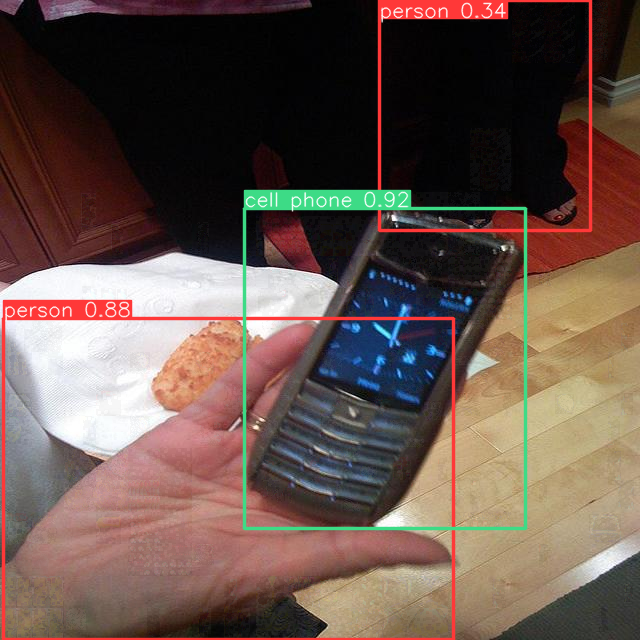}
        \caption{}
        \label{fig:atk_eval_1}
    \end{subfigure}
    \begin{subfigure}{.48\linewidth}
        \centering
        \includegraphics[width=1.0\linewidth]{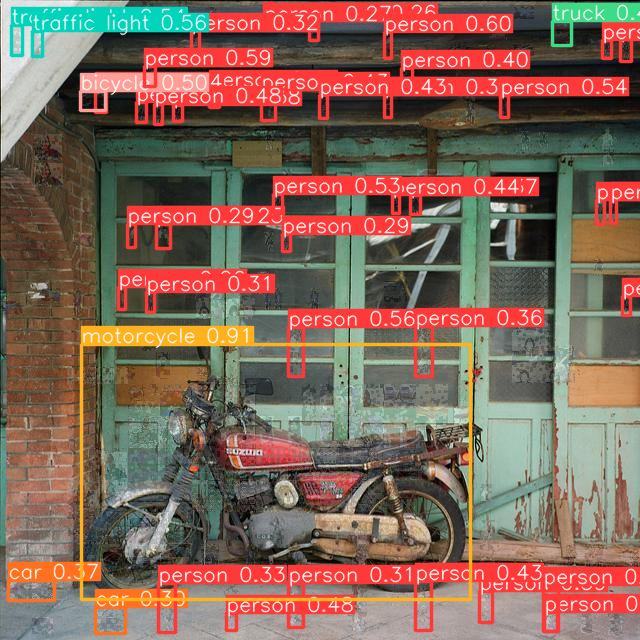}
        \caption{}
        \label{fig:atk_eval_2}
    \end{subfigure}
    \caption{The strength of adversarial examples highly depends on color fluctuations. 
    }
    \label{fig:atk_eval}
\end{minipage}
\begin{minipage}[t]{.48\linewidth}
\centering
    \begin{subfigure}{.48\linewidth}
        \centering
        \includegraphics[width=1.0\linewidth]{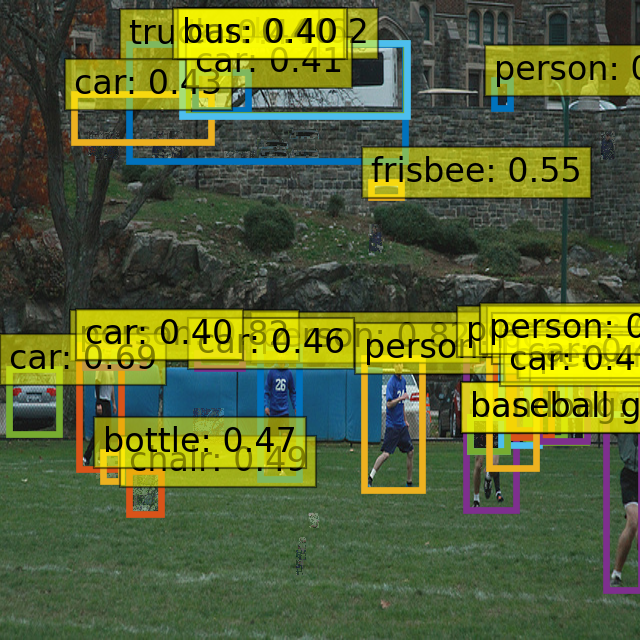}
        \caption{}
        \label{fig:atk_spatial_1}
    \end{subfigure}
    \begin{subfigure}{.48\linewidth}
        \centering
        \includegraphics[width=1.0\linewidth]{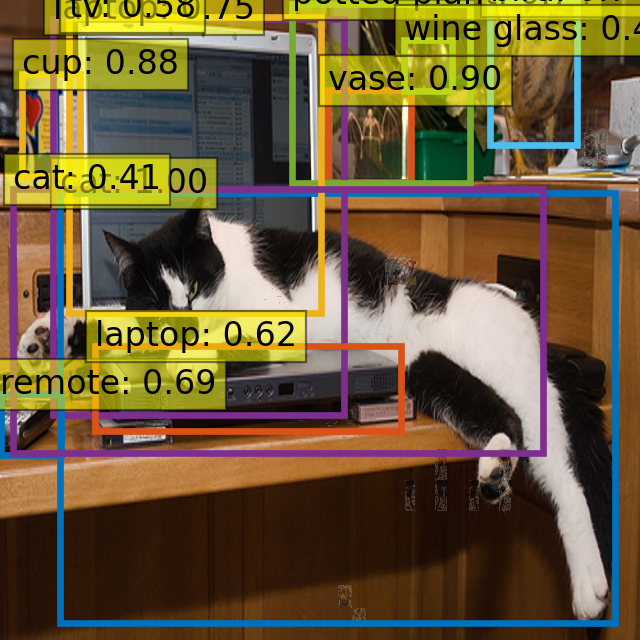}
        \caption{}
        \label{fig:atk_spatial_2}
    \end{subfigure}
    \caption{Objects tend to be clustered in specific regions.}
    \label{fig:atk_spatial}
\end{minipage}
\end{figure}

The experimental results shown in \Cref{table:ASR_model_eps} suggests that the data collected from different datasets have no significant impact on ASR. However, we would like to emphasize that the strength of adversarial attacks highly depends on the color distribution between selected target images and collected data. As shown in \Cref{fig:atk_eval_1}, the upper region of the left image is almost entirely black. It becomes challenging to match an object from the database that aligns with this color distribution. Conversely, \Cref{fig:atk_eval_2} exhibits wide color fluctuations across the entire spatial domain, providing attackers with the ability to embed ghost objects at any location.

\subsection{Ablation study}

\subsubsection{Analysis of spatial distribution}
Initially, we expected the position-centric algorithm to generate numerous objects uniformly distributed across the entire image. However, as depicted in \Cref{fig:atk_spatial}, the generated objects tend to cluster in specific regions, a pattern consistent with findings reported in \cite{chen2023overload}. When objects crowd into a small region, there is a high probability that these objects reference the same item. Consequently, numerous candidates might be eliminated during an internal stage. Conversely, empty regions may contain objects with relatively low confidence, leading the model to not report these objects to users. This clustering phenomenon presents a challenge in generating ghost objects. To tackle this issue, it is advisable to implement a spatial attention algorithm similar to the Overload attack or to identify specific areas that should be preserved. Besides, it is imperative to design an appropriate approach to hinder high-confidence objects.

\subsubsection{Minimum Number of Required Images}

This experiment aims to evaluate the influence of the number of collected images on ASR. We followed the same implementation details outlined in \Cref{sec:exp:setup}, the epsilon radius was set to 32, and the data were collected from MS COCO dataset. \Cref{table:ASR_num} presents the experimental results.

As can be seen, data diversity grows as the number of collected images increases, leading to an improvement in ASR. However, these improvements are not notably significant, primarily due to each dataset having its own distinct data distribution. For instance, there is an abundance of objects with green textures such as grasslands, baseball fields, and trees. Nevertheless, as illustrated in \Cref{fig:atk_eval_1}, these objects are incompatible with black backgrounds. Consequently, despite the increased number of collected images, the enhancements remain marginal. These experimental findings underscore the importance of data diversity. To enhance ASR, it's crucial to consider gathering data from diverse datasets or implementing color transformations.

\subsubsection{Influence on Using Different Datasets or Models}

\begin{table}[t]
    \begin{minipage}[t]{.43\linewidth}
        
        \caption{Ablation study for the number of images collected on ASR.}
        \centering
        \label{table:ASR_num}
        \begin{tabular}{c|c|c|c}
        \multirow{2}{*}{Model}   & \multicolumn{3}{|c}{Images [\#]} \\
         & 100 & 200 & 500 \\
        \hline
        DERT            & 29\% & 31\% & 32\% \\
        Faster-RCNN     & 35\% & 37\% & 37\% \\
        Retinanet       & 37\% & 39\% & 42\% \\
        FCOS            & 49\% & 52\% & 55\% \\
        YOLOv8          & 81\% & 83\% & 84\% \\
        \end{tabular}
    \end{minipage}
    \hfill
    \begin{minipage}[t]{.55\linewidth}
        \caption{Ablation study for configurations of data collection on ASR.}
        \centering
        \label{table:ASR_conf}
        \begin{tabular}{c|c|c|c|c}
        Model        & config 1 & config 2 & config 3 & config 4\\
        \hline
        DERT            & 29\% & 28\% & 29\% & 28\% \\
        Faster-RCNN     & 35\% & 36\% & 31\% & 36\% \\
        Retinanet       & 37\% & 35\% & 32\% & 39\% \\
        FCOS            & 49\% & 47\% & 43\% & 52\% \\
        YOLOv8          & 81\% & 81\% & 81\% & 83\% \\
        \end{tabular}
    \end{minipage}
\end{table}

This experiment aims to evaluate the configurations of data collection on ASR and the epsilon radius was set to 32. In the first configuration, patches were collected in advance from the MS COCO dataset using the target model. The second configuration followed the same approach but substituted the MS COCO dataset with the ImageNet dataset \cite{deng2009imagenet}. The third configuration involved collecting candidate objects from the MS COCO dataset using the YOLOv8 model. The last configuration leveraged the GCP vision API to collect patches. The experimental results are summarized in \Cref{table:ASR_conf}.

As can be seen, no significant performance gap among different configurations although the locations or sizes of objects predicted by different models are not precise. These experimental results show data collection is a crucial step, but it is not currently the performance bottleneck of this work. Besides, the cost of the proposed attacks is affordable, which motivates attackers to exploit vulnerabilities in AI systems and invest in refining attack algorithms.

\subsection{Time Consumption and Costs}

Based on the above analysis, we can conclude that approximately 500 images are required to collect the patches and only one model needs to be deployed locally. We estimated the time consumption and total costs for data collection under the above scenario.

The inference time for processing 500 images with Nvidia V100 is less than one hour. Considering the cost of renting the same GPU on GCP for one hour (\$2.48 dollars per GPU), the total expense, including the time for initialization, would not exceed \$3 dollars. In contrast, the cost of using the GCP vision API is \$1.5 dollars per 1,000 requests. However, employing the GCP Vision API involves formatting data packets and transferring data between the server and the client. As a result, a single request takes no more than 10 seconds, leading to a total data collection time of approximately 2 hours.

From the comprehensive analysis, it is unequivocally evident that deploying a private model locally as the most economically viable solution for the proposed algorithm.

\subsection{Discussion and Limitations}
Parallel works \cite{cai2022zero,cai2022context}  have suggested that a co-occurrence graph, which captures relationships between pairs of objects, can provide suggested placements for attaching these objects.  However, data collection remains an essential step. The main focus of these studies lies in minimizing the overall number of queries needed to create an adversarial example capable of deceiving the model. Besides, our approach distinguishes itself by offering a practical advantage in terms of incurring manageable costs associated with data collection.

Nonetheless, we observed that specific classes tend to appear frequently in the perturbed images. Moreover, with the GCP vision API, certain specific classes, like '2D barcode,' can suppress the appearance of other objects, even when the pairwise distances are much larger. This could lead to a failure in our attacking algorithm. These observations seem to suggest that the GCP Vision API employs multiple processing flows for certain conditions. Real-world AI applications are more complex than what we considered in this work, and it is absolutely necessary to collect more data to analyze victim model preferences in order to improve ASR. The time investment and total costs must exceed the results of this work.

Data collection with public AI systems is not recommended. GCP officially states that the size limits for input and output data are 20MB and 10MB, respectively. If the limit is exceeded, the system will reject the response. However, we encountered some exceptions during our experiments despite not violating these limits. For instance, frequent accesses may trigger bandwidth limitations, slowing down the speed of data transfer. Additionally, certain images may cause glitches, leading the system to directly reject the requests. This suggests that GCP has an internal system for managing user requests and available resources.


The proposed attack strategy has a significant limitation: it is inherently size-variant and cannot generate adversarial examples within a narrow epsilon ball. Specifically, the input images are resized to a pre-defined size before being fed into the target model. However, the exact dimensions remain undisclosed to end-users. In an extreme scenario, a system with limited computational resources may compress inputs to small dimensions, like $(64, 64)$, to reduce the computing costs. The compression invariably eliminates the fine structure of perturbations. Consequently, multiple adversarial examples produced during internal steps are mapped to the same resized image, rendering numerous queries invalid. Nevertheless, we believe that integrating Expectation over Transformation (EoT) \cite{athalye2018synthesizing,athalye2018obfuscated} into the proposed attack could enhance its effectiveness.

Besides, the strength of black-box attacks is much weaker than that of white-box attacks. According to the criteria set in this paper, the ASR of white-box attacks stands at 100\% but with single image, fewer iterations, and a smaller epsilon ball. However, we emphasis that the primary contributions of the study is exploiting
vulnerabilities in AI systems and analysing the feasibility of latency attacks on the real-world AI applications.

Those limitations and our observations implicitly propose three probable defenses. The first one performs multiple inferences with various input dimensions. If results are divergent significantly, there is a high probability that the input image is malicious. The second approach involves integrating a mechanism that can measure context inconsistency \cite{li2020connecting}. Requests are rebuffed if distinct and unrelated objects coexist within the same image. The third strategy revolves around image quality assessment. Since adversarial perturbations at the current stage live in relatively large epsilon ball, they might be detectable by some existing image quality measurements.

\section{Conclusion}
This paper has presented a novel attack strategy, the "steal now, attack later" approach, which aims to deceive the victim model into generating ghost objects under the black-box scenario such that the system cannot respond to the requests immediately or computing resources are exhausted. This is the first paper studies the feasibility of latency attacks against object detection under the black-box scenario. This innovative method resolves the issue of the lack of information about unqualified objects. The experimental results have demonstrated the effectiveness of this approach across various models. Furthermore, the proposed attack is shown to be applicable to public vision APIs provided by Google Cloud Platform (GCP). Through a comprehensive analysis, it has been determined that deploying a private model locally as the most economical solution, supported by affordable costs associated with the proposed attack. This encourages attackers to invest in improving attack algorithms to exploit vulnerabilities in AI systems.

%
%
\bibliographystyle{splncs04}
\bibliography{main}
\appendix

\section{Data Visualization}
\label{sec:data_visual}
\begin{figure}[!t]
\centering
    \begin{subfigure}{.49\linewidth}
        \centering
        \includegraphics[width=1.0\linewidth]{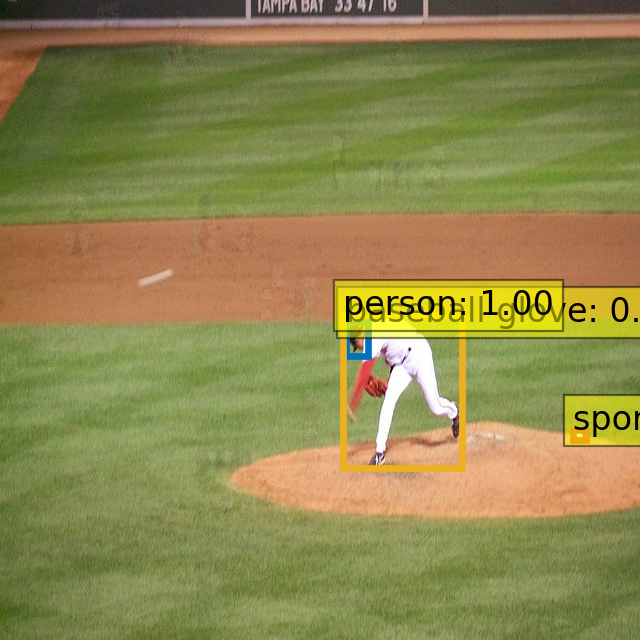}
        \caption{$\epsilon=8$}
        \label{fig:pred_eps_dert_example_1_eps_8}
    \end{subfigure}
    \begin{subfigure}{.49\linewidth}
        \centering
        \includegraphics[width=1.0\linewidth]{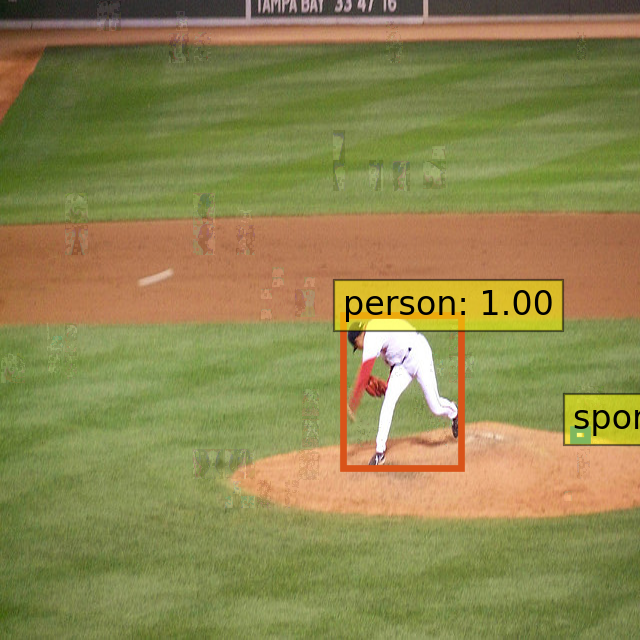}
        \caption{$\epsilon=16$}
        \label{fig:pred_eps_dert_example_1_eps_16}
    \end{subfigure}
    \begin{subfigure}{.49\linewidth}
        \centering
        \includegraphics[width=1.0\linewidth]{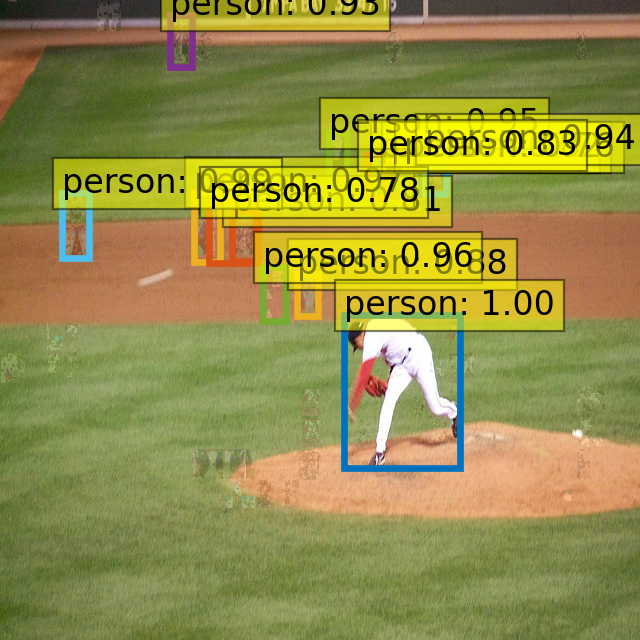}
        \caption{$\epsilon=24$}
        \label{fig:pred_eps_dert_example_1_eps_24}
    \end{subfigure}
    \begin{subfigure}{.49\linewidth}
        \centering
        \includegraphics[width=1.0\linewidth]{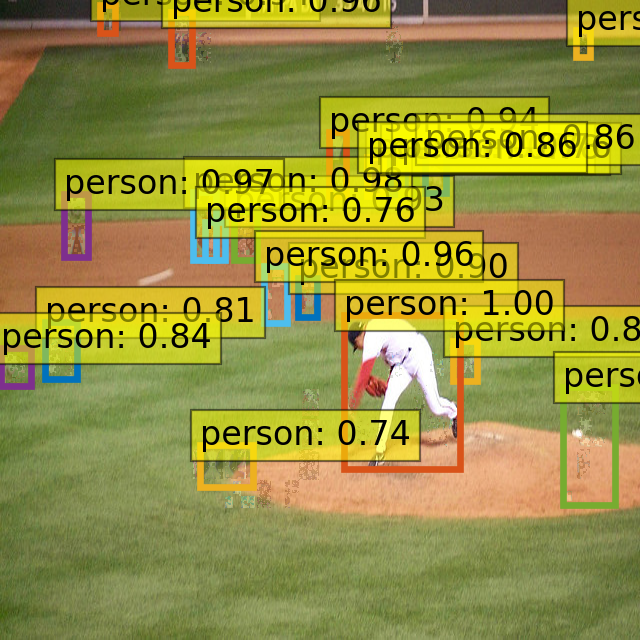}
        \caption{$\epsilon=32$}
        \label{fig:pred_eps_dert_example_32}
    \end{subfigure}
    \caption{The predictions of adversarial examples by DERT under differ epsilon radii.}
    \label{fig:pred_eps_dert_example_1}
\end{figure}

\begin{figure}[!t]
\centering
    \begin{subfigure}{.49\linewidth}
        \centering
        \includegraphics[width=1.0\linewidth]{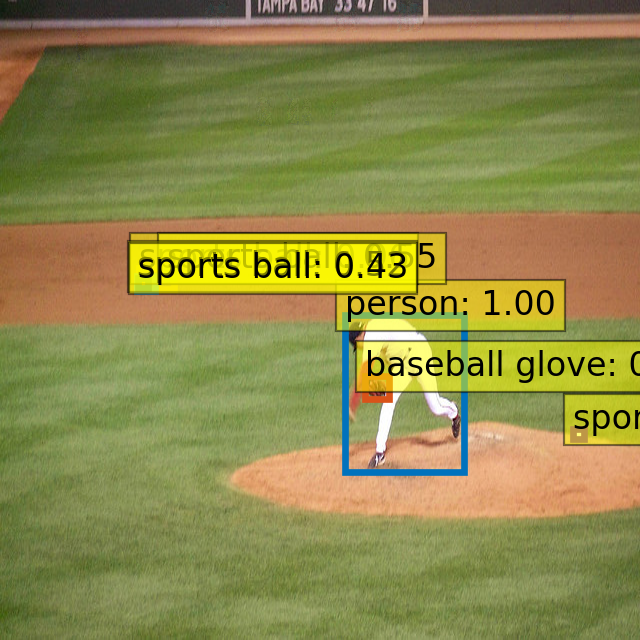}
        \caption{$\epsilon=8$}
        \label{fig:pred_eps_fasterrcnn_example_1_eps_8}
    \end{subfigure}
    \begin{subfigure}{.49\linewidth}
        \centering
        \includegraphics[width=1.0\linewidth]{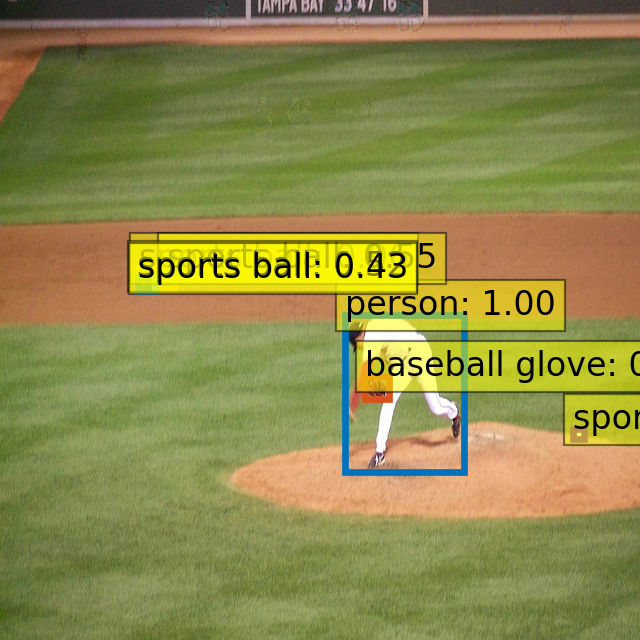}
        \caption{$\epsilon=16$}
        \label{fig:pred_eps_fasterrcnn_example_1_eps_16}
    \end{subfigure}
    \begin{subfigure}{.49\linewidth}
        \centering
        \includegraphics[width=1.0\linewidth]{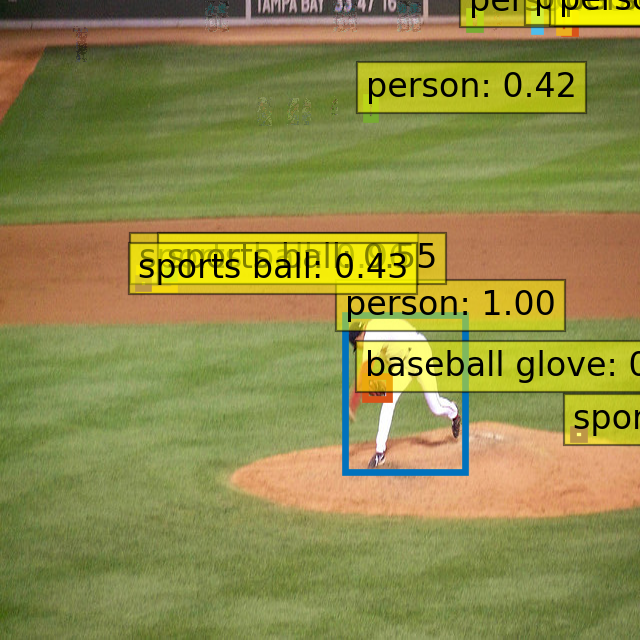}
        \caption{$\epsilon=24$}
        \label{fig:pred_eps_fasterrcnn_example_1_eps_24}
    \end{subfigure}
    \begin{subfigure}{.49\linewidth}
        \centering
        \includegraphics[width=1.0\linewidth]{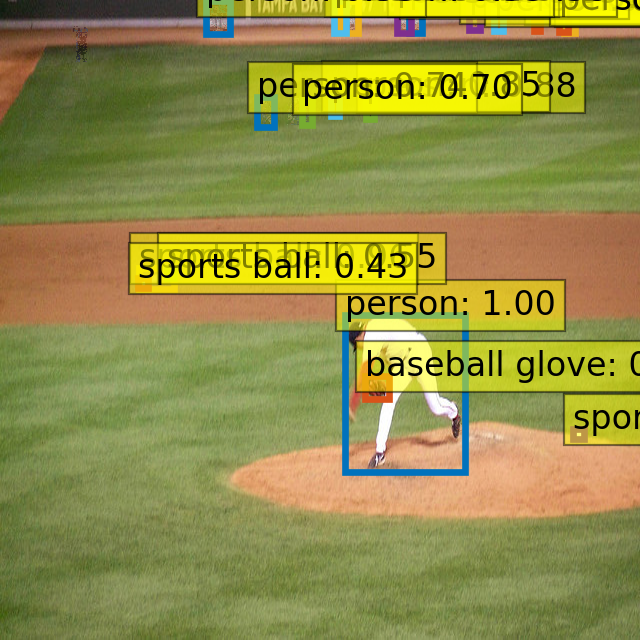}
        \caption{$\epsilon=32$}
        \label{fig:pred_eps_fasterrcnn_example_32}
    \end{subfigure}
    \caption{The predictions of adversarial examples by faster R-CNN under differ epsilon radii.}
    \label{fig:pred_eps_fasterrcnn_example_1}
\end{figure}

\begin{figure}[!t]
\centering
    \begin{subfigure}{.49\linewidth}
        \centering
        \includegraphics[width=1.0\linewidth]{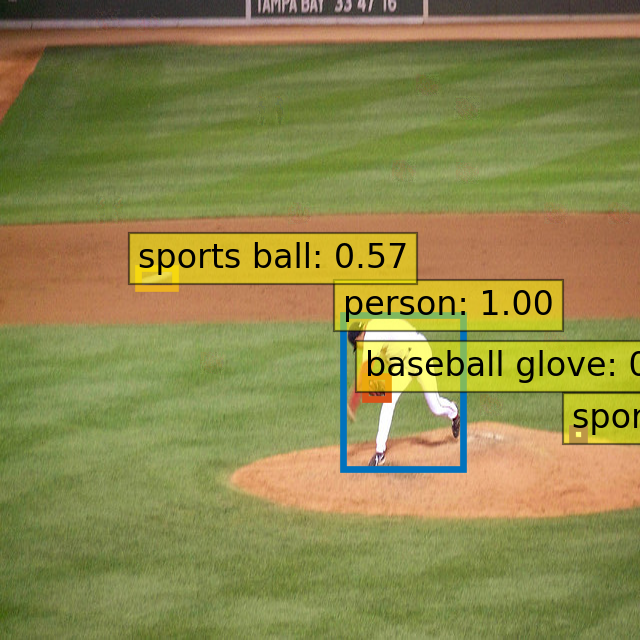}
        \caption{$\epsilon=8$}
        \label{fig:pred_eps_retinanet_example_1_eps_8}
    \end{subfigure}
    \begin{subfigure}{.49\linewidth}
        \centering
        \includegraphics[width=1.0\linewidth]{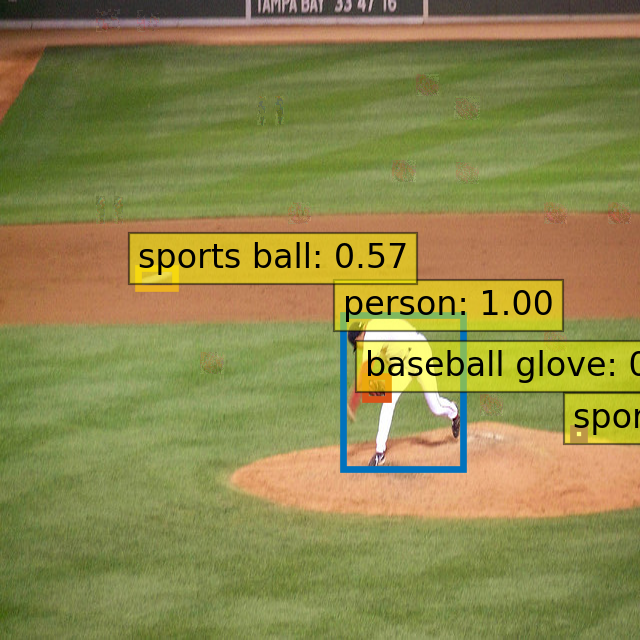}
        \caption{$\epsilon=16$}
        \label{fig:pred_eps_retinanet_example_1_eps_16}
    \end{subfigure}
    \begin{subfigure}{.49\linewidth}
        \centering
        \includegraphics[width=1.0\linewidth]{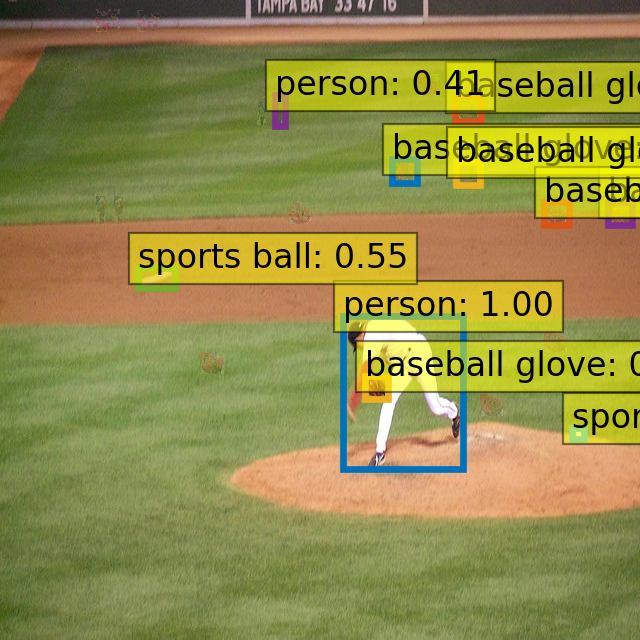}
        \caption{$\epsilon=24$}
        \label{fig:pred_eps_retinanet_example_1_eps_24}
    \end{subfigure}
    \begin{subfigure}{.49\linewidth}
        \centering
        \includegraphics[width=1.0\linewidth]{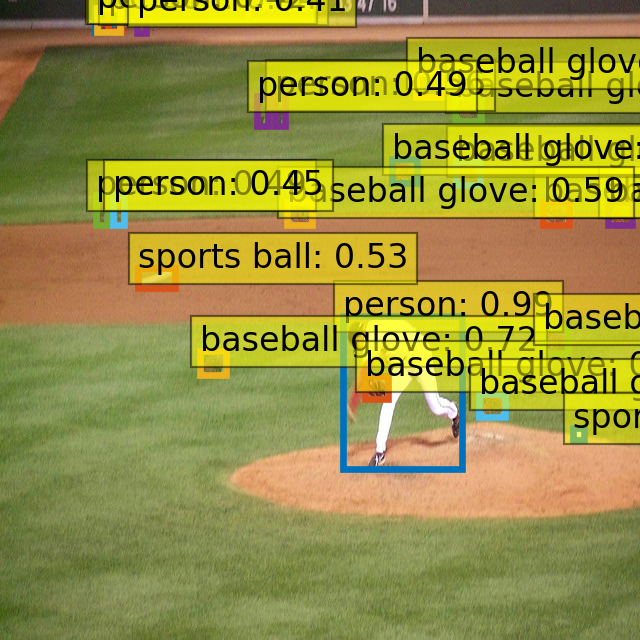}
        \caption{$\epsilon=32$}
        \label{fig:pred_eps_retinanet_example_32}
    \end{subfigure}
    \caption{The predictions of adversarial examples by Retinanet under differ epsilon radii.}
    \label{fig:pred_eps_retinanet_example_1}
\end{figure}

\begin{figure}[!t]
\centering
    \begin{subfigure}{.49\linewidth}
        \centering
        \includegraphics[width=1.0\linewidth]{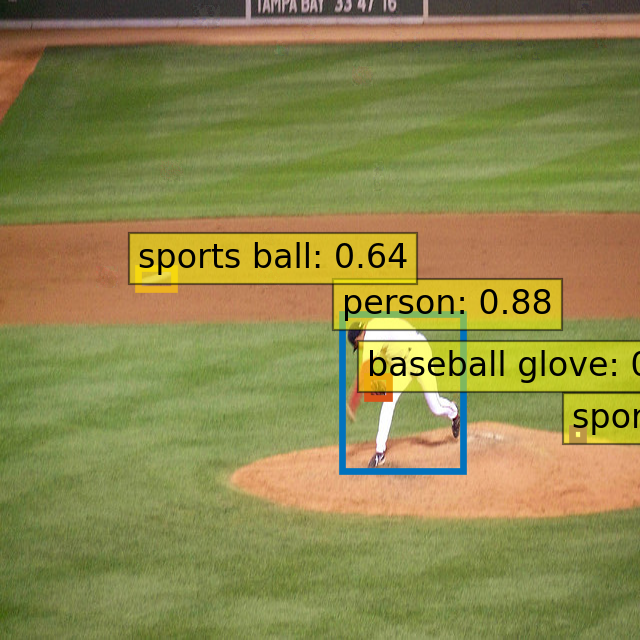}
        \caption{$\epsilon=8$}
        \label{fig:pred_eps_fcos_example_1_eps_8}
    \end{subfigure}
    \begin{subfigure}{.49\linewidth}
        \centering
        \includegraphics[width=1.0\linewidth]{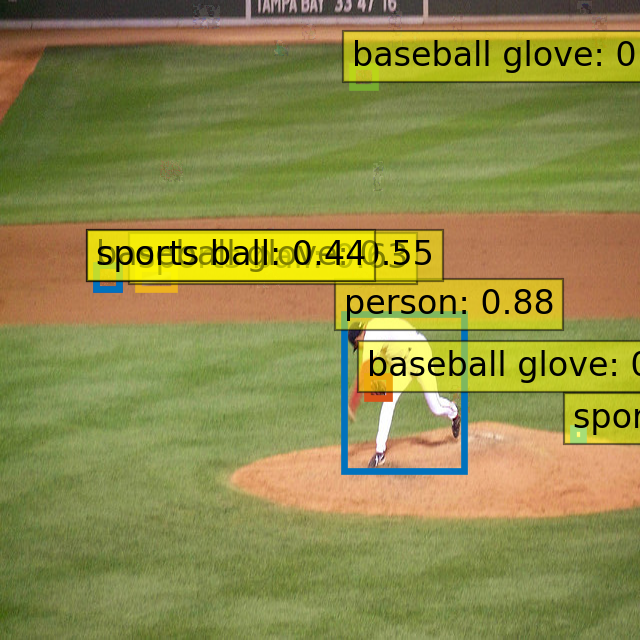}
        \caption{$\epsilon=16$}
        \label{fig:pred_eps_fcos_example_1_eps_16}
    \end{subfigure}
    \begin{subfigure}{.49\linewidth}
        \centering
        \includegraphics[width=1.0\linewidth]{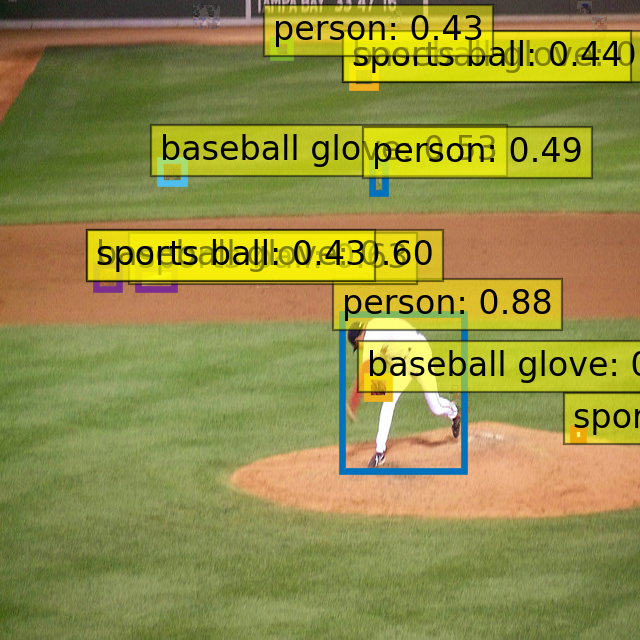}
        \caption{$\epsilon=24$}
        \label{fig:pred_eps_fcos_example_1_eps_24}
    \end{subfigure}
    \begin{subfigure}{.49\linewidth}
        \centering
        \includegraphics[width=1.0\linewidth]{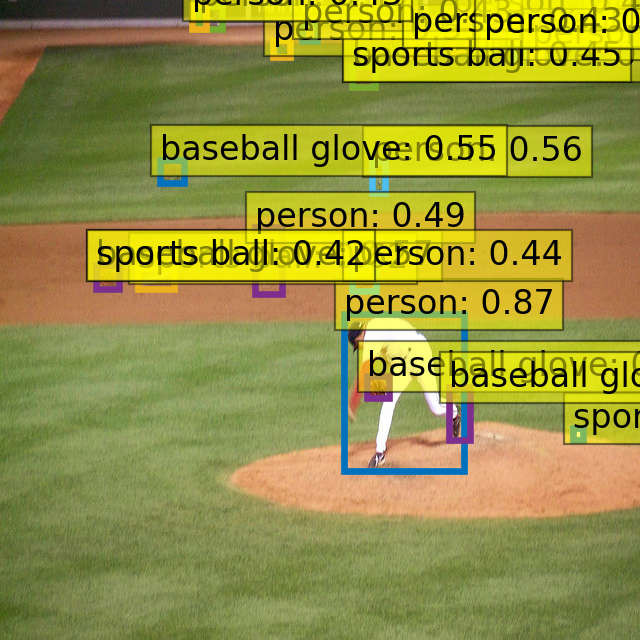}
        \caption{$\epsilon=32$}
        \label{fig:pred_eps_fcos_example_32}
    \end{subfigure}
    \caption{The predictions of adversarial examples by FCOS under differ epsilon radii.}
    \label{fig:pred_eps_fcos_example_1}
\end{figure}

\begin{figure}[!t]
\centering
    \begin{subfigure}{.49\linewidth}
        \centering
        \includegraphics[width=1.0\linewidth]{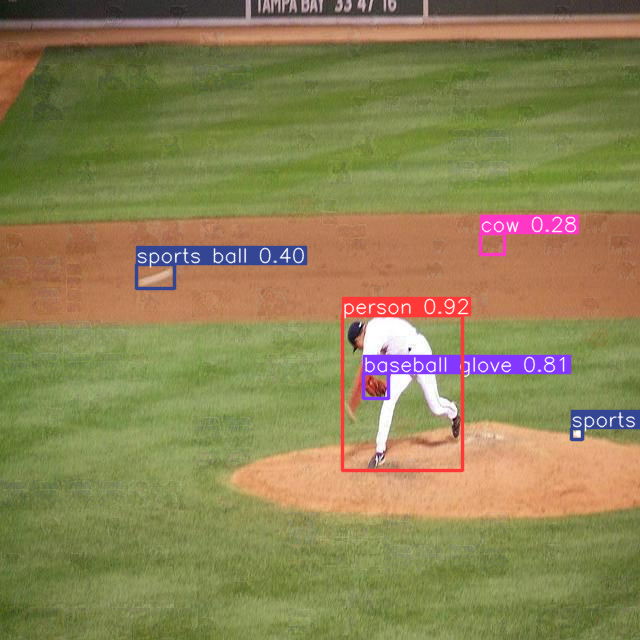}
        \caption{$\epsilon=8$}
        \label{fig:pred_eps_YOLOv8_example_1_eps_8}
    \end{subfigure}
    \begin{subfigure}{.49\linewidth}
        \centering
        \includegraphics[width=1.0\linewidth]{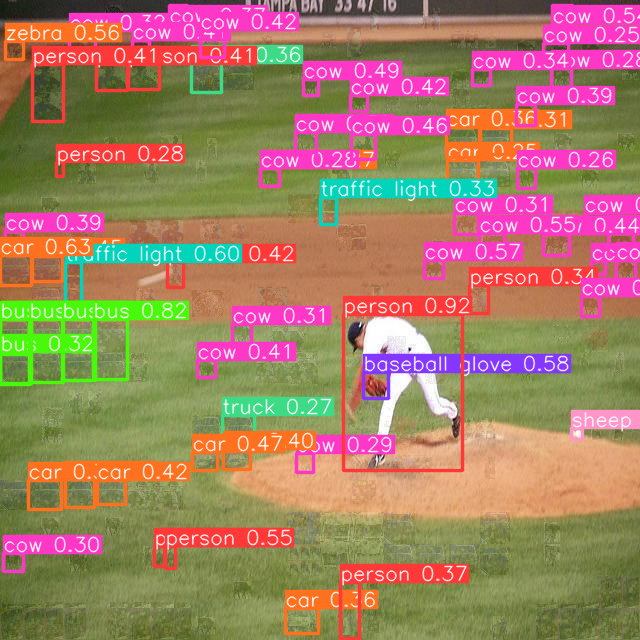}
        \caption{$\epsilon=16$}
        \label{fig:pred_eps_YOLOv8_example_1_eps_16}
    \end{subfigure}
    \begin{subfigure}{.49\linewidth}
        \centering
        \includegraphics[width=1.0\linewidth]{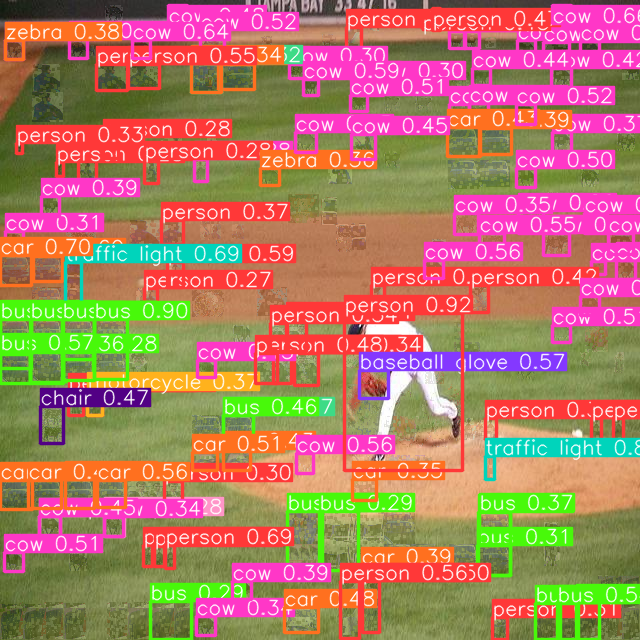}
        \caption{$\epsilon=24$}
        \label{fig:pred_eps_YOLOv8_example_1_eps_24}
    \end{subfigure}
    \begin{subfigure}{.49\linewidth}
        \centering
        \includegraphics[width=1.0\linewidth]{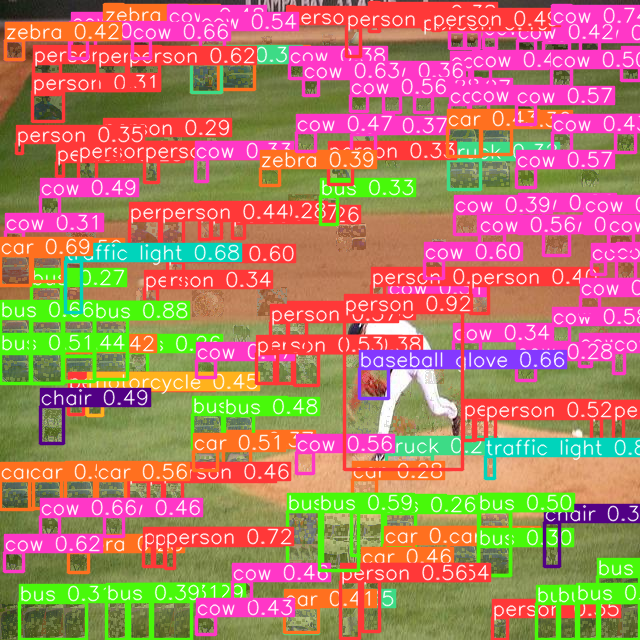}
        \caption{$\epsilon=32$}
        \label{fig:pred_eps_YOLOv8_example_32}
    \end{subfigure}
    \caption{The predictions of adversarial examples by YOLOv8 under differ epsilon radii.}
    \label{fig:pred_eps_YOLOv8_example_1}
\end{figure}

In this section, we present visualizations of the predictions made by various models for the adversarial examples, as shown in \Cref{fig:pred_eps_dert_example_1} to \Cref{fig:pred_eps_YOLOv8_example_1}. It is evident that all models have failed when subjected to $\epsilon=8$. Moreover, the objects generated by adversarial examples tend to be relatively small in size. According those observations, we believe that this is the main reason why our attack cannot be applied to SSD currently.

\end{document}